\documentclass[lettersize,journal]{IEEEtran}
\usepackage{amsmath,amsfonts}
\usepackage{algorithmic}
\usepackage{algorithm}
\usepackage{array}
\usepackage[caption=false,font=normalsize,labelfont=sf,textfont=sf]{subfig}
\usepackage{textcomp}
\usepackage{stfloats}
\usepackage{url}
\usepackage{verbatim}
\usepackage{graphicx}
\usepackage{xcolor}  
\usepackage{cite}
\usepackage{multirow}  
\usepackage{booktabs} 
\usepackage{lmodern}

\usepackage{multicol}
\usepackage[utf8]{inputenc}  

\usepackage{flushend}

\begin{document}
	
	\title{Edge Detection based on Channel Attention \\and Inter-region Independence Test}
	
	\author{Ru-yu Yan,  Da-Qing Zhang 
		\thanks{Corresponding author: Da-Qing Zhang(e-mail: d.q.zhang@ustl.edu.cn).}
		\thanks{The authors are with the Department of School of Science, University of Science and Technology Liaoning, Anshan 114051, PR China.}}
	
	
	\maketitle

	\markboth{Journal of \LaTeX\ Class Files,~Vol.~xxx, No.~x, month~year}%
	{Shell \MakeLowercase{\textit{et al.}}:Edge Detection based on Channel Attention and Inter-region Independence Test}

	\begin{abstract}
		Existing edge detection methods often suffer from noise amplification and excessive retention of non-salient details, limiting their applicability in high-precision industrial scenarios. To address these challenges, we propose CAM-EDIT, a novel framework that integrates Channel Attention Mechanism (CAM) and Edge Detection via Independence Testing (EDIT). The CAM module adaptively enhances discriminative edge features through multi-channel fusion, while the EDIT module employs region-wise statistical independence analysis (using Fisher's exact test and chi-square test) to suppress uncorrelated noise.Extensive experiments on BSDS500 and NYUDv2 datasets demonstrate state-of-the-art performance. Among the nine comparison algorithms, the F-measure scores of CAM-EDIT are 0.635 and 0.460, representing improvements of 19.2\% to 26.5\% over traditional methods (Canny, CannySR), and better than the latest learning based methods (TIP2020, MSCNGP). Noise robustness evaluations further reveal a 2.2\% PSNR improvement under Gaussian noise compared to baseline methods. Qualitative results exhibit cleaner edge maps with reduced artifacts, demonstrating its potential for high-precision industrial applications.
		
	\end{abstract}
	

\section{Introduction}

\IEEEPARstart{E}{dge} detection is a fundamental problem in image processing and computer vision, as it aiming to identify significant intensity changes in an image that typically correspond to object boundaries or texture discontinuities. Accurate edge detection plays a vital role in numerous applications, including medical imaging\cite{di_jing_medical_2023}, autonomous driving\cite{Barry2019xYOLOAM}, aerospace and remote sensing \cite{ma_weixuan_edge_2023}, facial recognition\cite{Lee17052024,9599570}, and image classification\cite{8110709}.

The correlation between adjacent pixels is a key factor in the design of robust edge detection algorithms. Traditional methods often struggle to achieve high accuracy and robustness, especially in the presence of noise, which can lead to false detections or missing crucial details. Addressing this limitation requires algorithms that are capable of distinguishing genuine edges from noise while preserving structural integrity.

This paper proposes a novel edge detection method that combines the Channel Attention Mechanism (CAM) with Edge Detection based on Independence Test (EDIT). The CAM module enhances edge features and reduces computational complexity by focusing on salient regions, while the EDIT module refines the results by evaluating statistical independence among pixel regions. Specifically, independence test, a statistical method used to assess whether two variables are independent, is introduced to partition the preliminary edge image into small regions and apply Fisher’s exact test and chi-square test. This approach effectively suppresses noise and retains correlated regions, identifying pixels within these regions as edge pixels.

Compared to existing algorithms, the proposed method demonstrates superior accuracy and noise resistance. In practical scenarios, adjacent pixels often share similar characteristics due to spatial continuity, image structure, and sensor properties\cite{Chan2002ACI}. For instance, variations in lighting and surface reflections can affect the intensity distribution of neighboring pixels\cite{10021826}. The proposed method leverages this intrinsic correlation to enhance edge detection performance and improve robustness. Despite decades of research, two persistent challenges remain:

	1.	Noise Sensitivity: Fixed-threshold gradient operators (e.g., Sobel, Canny) amplify high-frequency noise.
	
	2.	Over-Segmentation: Attention mechanisms like SE Block emphasize global features while neglecting local edge continuity.
	
To address these issues, we propose CAM-EDIT with three key innovations:

	1.	Channel Attention for Edge Enhancement: CAM dynamically highlights discriminative edge features across RGB channels, reducing background interference. The attention mechanism can dynamically adjust the model's attention to different regions, making it more focused on the true edges rather than background noise or texture interference.
	
	2.	We use the Sigmoid membership function instead of linear normalization. This nonlinear mapping smoothes eigenvalues and improves the performance of tasks such as edge detection and fuzzy logic systems. In these tasks, smooth and continuous conversion is crucial. 
	
	3.	Statistical Independence Test: EDIT partitions images into regions and applies Fisher's exact and chi-square tests to retain spatially correlated edges, effectively filtering random noise.

\newpage

	\IEEEpubidadjcol
	
\subsubsection*{\bf Experiments on BSDS500 show CAM-EDIT achieves F1=0.635, which was 26.5\% higher than that of CannySR. And experiments on NYUDv2 show CAM-EDIT achieves F1=0.460, which is 19.2\% higher than that of CannySR. Contributions include}
\begin{itemize}
	\item{A parameter-efficient CAM architecture for multi-scale edge feature fusion.}
	
	\item{A statistically-driven EDIT strategy for noise-resistant edge verification.}
	
	\item{Comprehensive validation on noise-corrupted datasets, demonstrating industrial applicability.}
\end{itemize}

The remainder of this paper is organized as follows: Section II reviews the progress of edge detection, as well as some related works to this study. Section III presents implementation details of the proposed CAM-EDIT. Section IV contains rigorous experiments and evaluation metrics analysis, and we summarize the whole paper in section V.

\section{Related Work}
\subsection{Traditional edge detection}

Early edge detection methods relied on manual extraction, which was time-consuming and unsuitable for processing complex images. In 1963, Roberts\cite{1998Edge} proposed the Roberts operator, which utilizes a local difference operator to approximate image gradients and locate contours. To improve detection accuracy, various gradient-based operators were subsequently developed, such as Sobel\cite{2003Pattern}, Prewitt, and Canny\cite{Canny1986A}. CannySR\cite{Akinlar2015} integrates Canny’s approach with ED’s routing step to generate structured edge segments. ED\cite{Cihan2012Edge} extracts thin, continuous edges by linking anchors along the gradient direction, while EDPF\cite{2012EDPF} applies the Helmholtz principle for fast, parameter-free edge extraction. Liu et al.\cite{Liu2020} introduced an adaptive thresholding method based on the linear correlation between 2D entropy and edge proportion. While these operators are effective for detecting edges, they are highly sensitive to noise, leading to discontinuities in edge maps\cite{Dhimish2022,Otamendi2023,Sun2020}. 

To mitigate noise, second-order derivative-based methods such as the Laplacian of Gaussian (LoG)\cite{2009Fast} were introduced. The performance of LoG depends heavily on the shape and size of the Gaussian filter. Later, the Difference of Gaussian (DoG)\cite{Dr2013Difference} and its extension, the XDoG algorithm, were proposed to achieve stronger noise resistance by convolving images with Gaussian functions. Despite these improvements, traditional operators only perform edge detection at a single scale, limiting their ability to detect local variations in complex images. Guo et al.\cite{fpgacanny} proposed that adaptive threshold technology outperforms traditional Canny in various lighting situations. Additionally, they often rely on fixed thresholds or parameters, resulting in incomplete edge curves.

Fuzzy logic effectively models nonlinear systems using flexible IF-THEN rules\cite{ijsrset}. It allows easy modifications without strict constraints, unlike knowledge-based methods requiring training. As an extension of multi-valued logic, it handles imprecise data and detects edges of varying thickness\cite{miedbsc}.

\subsection{Methods of deep learning}
With the advancement of machine learning, genetic algorithms (GAs) and their variants have been widely applied to edge detection. Artificial neural networks (ANNs) and deep learning techniques analyze images to effectively detect edges, mimicking the human nervous system by leveraging high-performance, low-level computational units. As flexible soft computing frameworks, these models handle various nonlinear problems, with processing units referred to as "neurons." Their structure is shaped by the characteristics of the input data\cite{ijsrset}.  

Several studies have explored GA-based improvements in edge detection. Abdel-Gawad et al. \cite{9143122} enhanced edge detection using GA, achieving an impressive 99.09\% accuracy in brain tumor detection, surpassing classical, fractional, and threshold-optimized methods. Kong et al. \cite{Kong2023} introduced an adaptive threshold Sobel operator based on GA, which enhances edge continuity and improves detection performance compared to conventional Sobel. Xia et al.\cite{xia2022novel} proposed a multi-scale edge detection approach using a hybrid wavelet transform. Their method is simple yet effective, outperforming Robert, Sobel, CNN, and multi-scale techniques employing B-spline or Gaussian wavelets while offering superior noise resistance and edge continuity. Hu et al. \cite{hu2022distance} developed an advanced edge detection model that surpasses earlier methods, generating high-quality edge images along with a corresponding distance field map.

Although these methods improve accuracy, they suffer from high computational complexity and are prone to local optima. Neural network-based methods\cite {2016Richer, 2021RHN} offer strong generalization ability and higher accuracy, but they require extensive training data, making them computationally expensive and memory-intensive.


\begin{figure}[]
	\centering
	\includegraphics[width=3.5in]{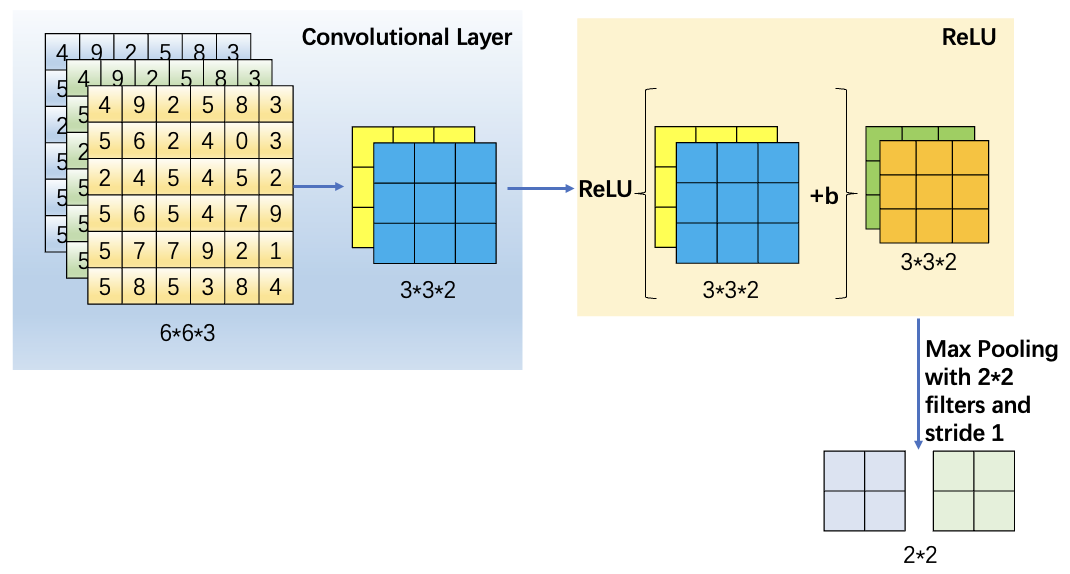}
	\caption{Overview of the proposed CAM scheme.}
	\label{channel attention}
\end{figure}

\section{Method}
This section will provide a detailed introduction to the proposed Channel Attention Mechanism-Edge Detection based on Independence Test method(CAM-EDIT), which is mainly divided into three steps: (1) Attention mechanism extracting edge features and calculates gradients; (2) coarse graining gradient images and morphological processing of binary images using dilation and erosion operations; (3) dividing the image into different regions and performing edge detection based on independence testing.

\subsection{Channel Attention-Based Feature Extraction}

We apply a channel attention mechanism to extract edge features from the three-channel input image. This mechanism assigns varying attention to each channel, enhancing feature importance and discriminability for precise edge extraction. The CAM architecture (Fig. \ref{channel attention}) includes a Dual-Convolution Layer, ReLU Activation, and a Pooling Layer.

\subsubsection{Dual-Convolution Layer}
\begin{figure}[]
	\centering
	\includegraphics[width=3in]{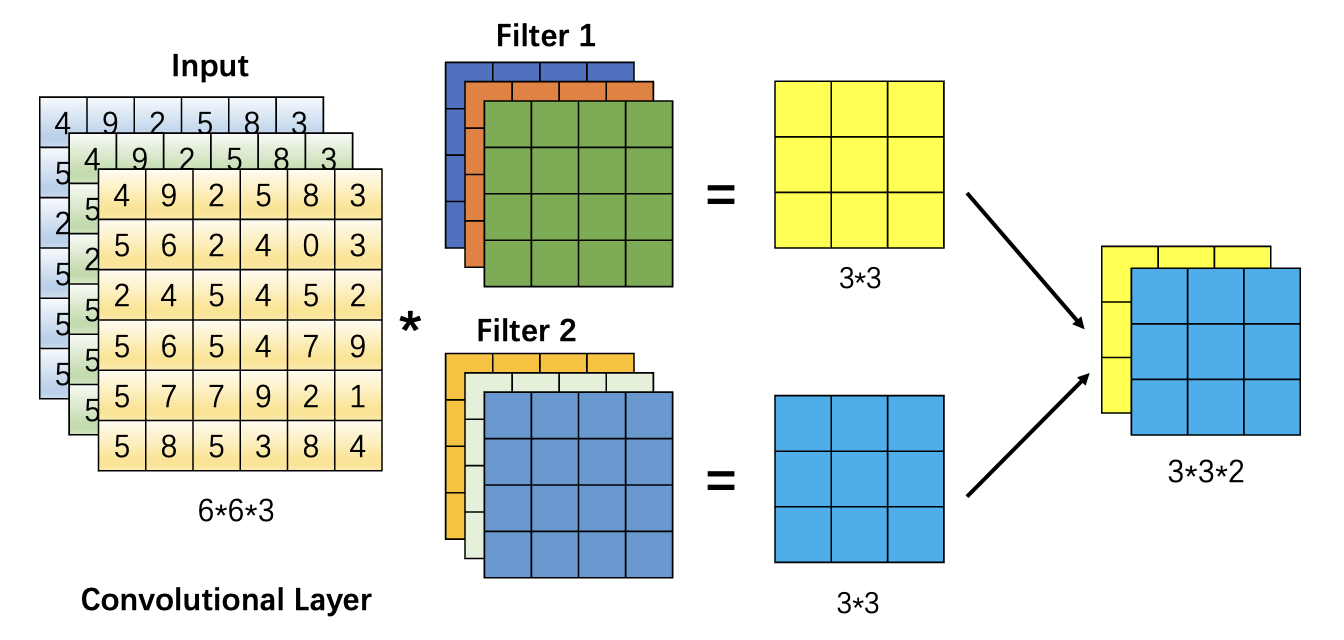}
	\caption{Schematic diagram of convolution operation.}
	\label{conv}
\end{figure}

This layer includes depthwise and pointwise convolutions. Depthwise convolution ($3 \times 3$) extracts local details per channel, while pointwise convolution ($2\times 2$) mixes channel information for broader feature capture, as shown in Fig. \ref{conv}. Convolution efficiently extracts features by sliding kernels over the image, leveraging parameter sharing and sparse interactions.  $3\times 3$ deep convolution is used to extract local features, and many classic networks (such as VGG\cite{10478014}, ResNet\cite{zhao2020deep}, MobileNet\cite{howard2017mobilenets}) use $3\times 3$ as the base convolution kernel. $2\times 2$ pointwise convolution is used for efficient fusion of channel information while controlling computational complexity. This combination can achieve a good balance between accuracy and computational complexity.

\subsubsection{ReLU Activation}
In order to enable the model to capture complex non-linear functional relationships, it is necessary to use activation functions to perform nonlinear transformations on linear inputs. We use the ReLU activation function for the process of non-linear transformation. The ReLU activation function is as follows:
	\begin{equation}
		f(x)=\left\{\begin{matrix}  0&x\le 0 \\  x&x>0\end{matrix}\right.
		\label{relu}
	\end{equation}
	
The ReLU function has the characteristic of sparse activation, which significantly improves the generalization ability and computational efficiency of the model, reduces unnecessary computational overhead, and effectively addresses the problem of gradient vanishing.

\subsubsection{Pooling Layer}
The pooling layer reduces feature dimensionality while preserving key information. It filters out important edges and ignores non-edge details. We use maximum pooling\cite{DESOUZABRITO2021115403} with a $2 \times 2$ window and a stride of 2, selecting the maximum value in each region to enhance feature representation.

Assuming the size of the input feature map $X$ is $m \times m $, after a max pooling operation of $n \times n $ size, the output feature map $Y$ is rounded by the ratio of $m$ to $n$ and can be denoted as $\left [ \frac{m}{n}  \right ] \times \left [ \frac{m}{n}  \right ] $. The mathematical formula for the max pooling operation can be expressed as:
	\begin{equation}
		Y_{i,j}=\text{max}(X_{ni:ni+n,nj:nj+n} )
		\label{pooling}
	\end{equation}
	where $Y_{i, j} $ is the element of the $i_{th} $ row and $j_{th} $ column of the output feature map $Y$. $X_ {ni: ni+n, nj: nj+n} $ is the slice of the pooling window in the input feature map $X$, starting at ($ni $, $nj $) and with a size of $n \times n $. Max represents taking the maximum value in the slice. 

Since the model focuses on edges, it remains robust even with noise or interference. To compute the attention weight for each channel, we apply a normalization process to aggregate spatial information across the feature map. The attention weight $\alpha _{c} $ for channel $c$ is computed as:
\begin{equation}
	\alpha _{c} =\sigma \left (   \frac{1}{HW} \sum_{i=1}^{H} \sum_{j=1}^{W} F_{c}(i,j)  \right )
	\label{weight}
\end{equation}
where $\sigma \left ( \cdot \right )$ denotes sigmoid activation.

\subsection{Preliminary Edge Detection Algorithm}

\begin{figure*}[]
	\centering
	\includegraphics[width=6in]{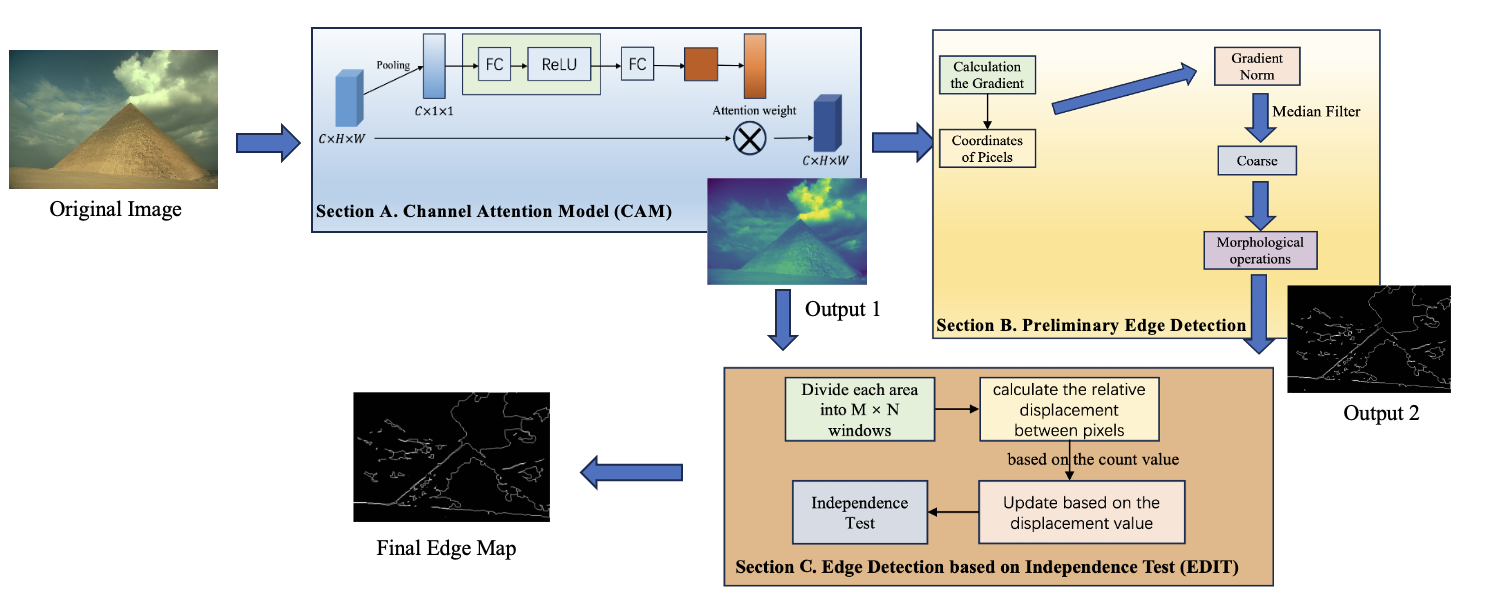}
	\caption{Overview of the proposed CAM-EDIT edge detection framework.}
	\label{framework}
\end{figure*}

\subsubsection{Calculating Gradients and Membership Degrees}
As discussed earlier, edge detection identifies significant local pixel changes. By computing image gradients, edges and textures can be detected. We use the Sobel operator, which applies $3 \times 3 $  kernels in horizontal ($G_{x} $) and vertical ($G_{y} $) directions to extract corresponding edges. Convolution is performed between these kernels and the image pixels to obtain $G_{x} $ and $G_{y} $, as shown in the following formula:

	\begin{equation}
		G_{x} =\begin{pmatrix}  -1&0  &1 \\  -2& 0 &2 \\  -1& 0 &1\end{pmatrix}\ast  A
		\label{gx}
	\end{equation}
	\begin{equation}
		G_{y} =\begin{pmatrix}  1 & 2  &1 \\  0& 0 &0 \\  -1& -2 & -1\end{pmatrix}\ast  A
		\label{gy}
	\end{equation}
	
Based on the above results, the grayscale value is calculated for each pixel in the image, resulting in a matrix $G$ of the same size as the image $A$: 
	\begin{equation}
		G=\sqrt{G_{x}^{2}+G_{y}^{2}  } 
		\label{g}
	\end{equation}
	
Then the gradient direction can be calculated  by using the following formula:
	\begin{equation}
		\Theta =argtan(\frac{G_{y} }{G_{_{x} } } )
		\label{theta}
	\end{equation}
	
To standardize pixel values, reduce bias, and improve efficiency, traditional normalization maps intensities to [0,1]. However, edge regions often show nonlinear transitions due to gradual brightness or color changes\cite{XIE20251291}, making linear models ineffective.  In fuzzy logic theory, membership functions provide a principled framework to characterize such nonlinear edge transitions\cite{kutlu_integrating_2024}, offering advantages in handling ambiguous bound aries. Thus, we replace linear normalization with a gradient-based membership function to maintain edge continuity while mapping values nonlinearly to [0,1].

Within fuzzy set theory, gradient-type membership functions (or smooth membership functions) quantify the degree of element affiliation to fuzzy sets through continuous, differentiable mappings\cite{de_literature_2022}. These functions are particularly effective in modeling transitional states between "membership" and "non-membership," a critical requirement for edge characterization. The Sigmoid membership function, as formulated in Equation \ref{fuzzy}, demonstrates intrinsic compatibility with nonlinear edge transitions. Its monotonic sigmoidal curve assigns higher membership degrees to regions with pronounced gradient magnitudes\cite{982883, low2020}, while suppressing low-contrast artifacts. This property aligns with the physical nature of edge formation, where gradient intensities correlate with boundary saliency: 

\begin{equation}
	\mu (x)=\frac{1}{1+e^{-k(x-x_{0} )} } 
	\label{fuzzy}
\end{equation}
where $k$ governs the transition steepness and $x_{0}$ denotes the inflection point. $x_{0}$ typically set as the mean or median of edge gradients. 

This formulation not only preserves edge continuity but also enhances robustness against quantization noise.A smaller $k$ results in a smoother Sigmoid transition, while a larger $k$ makes it steeper. In this paper, we define the pixel gradient value as $x$, the median gradient value as $x_{0}$, and set $k=5$.

\subsubsection{Median Filter Coarsening and Morphological Smoothing of Images}

Pixel correlation is crucial in edge detection, as it helps accurately locate edges. Convolutional filters (e.g., mean\cite{hu_jinshan_improved_2021}, median\cite{sambamurthy_scalable_2024}, and Gaussian filters\cite{chen_yun_optimization_2021}) enhance edge features by smoothing images and computing gradients. We use a median filter for noise reduction, as it effectively removes noise while preserving edges.

Median filtering is a nonlinear smoothing method that replaces each pixel with the median of its surrounding val ues, reducing noise while maintaining true pixel intensity. Coarse graining is essential for high-quality image pro cessing, such as in chemically\cite{yu_chemically_2024}, as it balances noise re duction and edge preservation. Morphological operations, including dilation and erosion, help refine image structures while maintaining edges.

Morphological operations, including dilation and erosion, further refine edges. Dilation expands objects by replacing pixels with the highest nearby value, filling gaps and connecting objects. In contrast, erosion shrinks objects by replacing pixels with the lowest nearby value, helping separate connected regions and contract edges.  
﻿

In this algorithm, we select a core size of $3 \times 3$ for expansion and corrosion operations. Due to the fact that this size only involves a small neighborhood, it can preserve the local details of the object. In addition, using a $3 \times 3$ kernel involves fewer pixels and can also reduce computational complexity.

\subsection{Edge Detection Method Based on Independence Test}

\begin{figure*}[]
	\centering
	\includegraphics[width=5.3in]{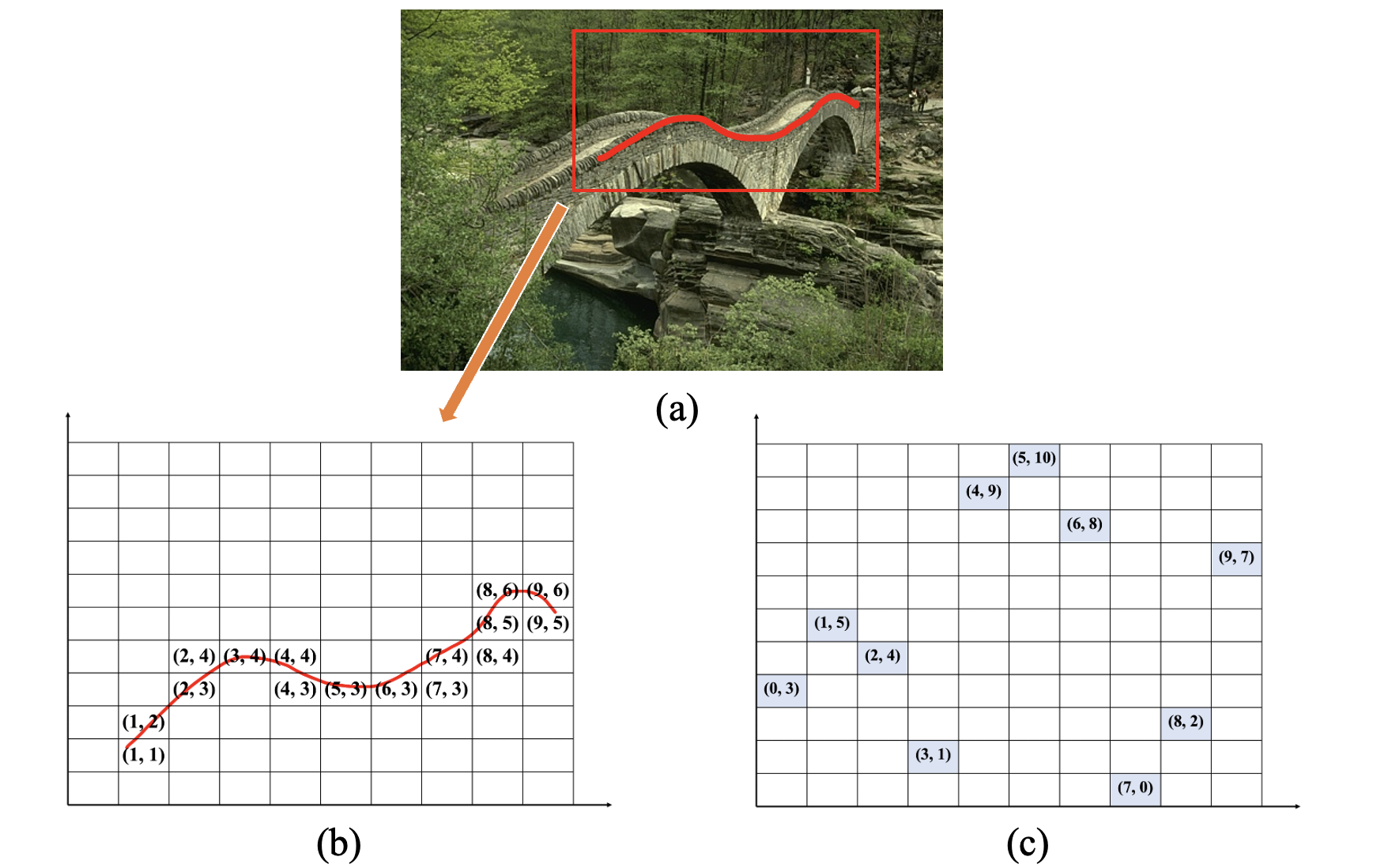}
	\caption{Extract continuous edge curves and randomly generated discrete points that can be regarded as noise points from the image.}
	\label{dlexample}
\end{figure*}

After the operation, we get a smooth edge image but with some detailed information and noise. Since image edges are usually continuous, we designed a method (EDIT) to divide the image into regions and check pixel coordinates for independence, as shown in Fig. \ref{framework}. This method keeps pixels with consistent coordinate changes as the final edge points.
	
Independence test is a testing method in statistics\cite{aaron2017}. It determines whether two categorical variables are independent of each other based on the frequency of events occurring. In this context, we define the set of edge pixel coordinates in the obtained edge image as $D$. The $x$ and $y$ coordinates of a pixel are the two categorical variables for independence testing. To determine the relationship between the coordinates of different pixels in the set $D$, we need to calculate the relative displacement of pixels along the horizontal and vertical axes, denoted as $\left |  \Delta x \right |$ and $\left |  \Delta y \right |$. Specifically, given a point $A(x_ {1}, y_ {1})$ in the set $D$, it is necessary to traverse all remaining points in the set $D$. 

And then calculate the relative displacement in two directions. Assuming that when traversing to point $B(x_ {2}, y_ {2})$ in set $D$, the relative displacement between point $A$ and point $B$ in the horizontal axis direction is $| \Delta x |=| x_ {1}-x_ {2} |$, and the relative displacement in the vertical axis direction is $| \Delta y |=| y_ {1}-y_ {2} |$. In order to easily count the frequency of different relative displacements, the relative displacement limit is set to $k$. Therefore, in the independence test, the value ranges of the two variables are respectively  divided into $\left \{ |\Delta x|<=k, |\Delta x|>k \right \} $ and $\left \{ |\Delta y|<=k, |\Delta y|>k \right \} $, and their coordinate point frequency sequence is shown in Table \ref{bzllb}. Among them, the frequency $a$ is the count of occurrences when $|\Delta x|<=k$ and $|\Delta y|<=k$. This means that there are a total of $a$ times in the coordinates of all points in the set $D$, where the relative displacement in both the horizontal and vertical directions is less than or equal to $k$. The frequencies $b$, $c$, and $d$ in the table are similarly.

\begin{table}[h]
	\centering
	\caption{Frequency contingency table of relative displacement among coordinates of different pixels in an edge image}
	\label{bzllb}
	\begin{tabular}{c  c c c }
		\toprule 
		& $\left |  \Delta y \right |\le k$ & $\left |  \Delta y \right |> k$ & total\\
		\midrule 
		$\left |  \Delta x \right |\le  k$ & $a$ & $b$  & $a+b$\\
		$\left |  \Delta x \right |> k$ & $c$ & $d$ & $c+d$ \\
		total & $a+c$ & $b+d$ & $n$ \\
		\bottomrule 
	\end{tabular}
\end{table}

Let's define the null hypothesis ($H_ {0}$) and the alternative hypothesis ($H_ {1}$) for the independence test for the categorical variables $x$ and $y$ coordinates:
\begin{list}{}{}
	\item{$H_ {0} $: $x$ coordinates and $y$ coordinates are independent of each other}
	\item{$H_ {1} $: $x$ coordinate and $y$ coordinate are not independent of each other, indicating a correlation between the coordinates.}
\end{list}

The frequencies in the contingency Table \ref{bzllb} can be divided into two situations: All frequencies are greater than 5, or there is at least one frequency less than 5. We adopt different methods for different situations: chi-square test and Fisher's exact test. They are all methods used to test whether there is independence between two categorical variables.

When the frequencies in the contingency table are all greater than 5, we use the chi-square test method. Assuming $H_ {0}$ is present, we calculate the chi-square statistic using the observed frequency and expected frequency. The formula is as follows:
	\begin{equation}
		\chi ^{2}=\sum \frac{(O_{ij}-E_{ij}  )^{2} }{E_{ij}}   
		\label{kafang}
	\end{equation}
	where $O_{ij}$ is the observed frequency, and $E_{ij}$ is the expected frequency. It represents the degree of deviation between observed values and theoretical values. Based on $\chi ^{2} $ distribution and the degree of freedom, it can be determined that the probability of obtaining the current statistic under the null hypothesis $H_ {0}$ is $p$. If the value of $p$ is less than the significance level (usually set to 0.05), we reject the null hypothesis $H_ {0}$ and accept the alternative hypothesis $H_ {1}$. This shows that the two coordinate variables are not independent of each other. It can also be said that the coordinate changes of the set $D$ are regular. Therefore,  these pixels can form the edges of the image. Otherwise, we accept the null hypothesis $H_{0}$ and consider the two coordinate variables to be independent of each other. It can also be said that the variation between the two coordinates of the points is irregular. So these points do not make up the edges of the image.

When the frequency is less than 5 in the contingency table, we use the Fisher's exact test method. Fisher's exact test, based on the idea of hypergeometric distribution. It describes the process of randomly selecting $n$ objects from a finite number of objects, and successfully extracting a specified type of object a certain number of times (without replacement). Fisher's exact test usually uses a $2\times 2$ contingency table, which is suitable for the problem of processing binary images in this article. The calculation formula for Fisher's exact test is as follows:
	\begin{equation}
		p=\frac{\begin{pmatrix}a+b \\a\end{pmatrix}\begin{pmatrix}c+d \\c\end{pmatrix}}{\begin{pmatrix}n \\a+c\end{pmatrix}} =\frac{\begin{pmatrix}a+b \\b\end{pmatrix}\begin{pmatrix}c+d \\d\end{pmatrix}}{\begin{pmatrix}n \\b+d\end{pmatrix}}
		\label{fisherbz}
	\end{equation}

The above formula calculates the probability of the distribution of the current $2\times 2$ contingency table occurring in all possible scenarios as $p$. If the value of $p$ is less than the significance level ($\alpha=0.05$), we reject the null hypothesis $H_ {0}$ and accept the alternative hypothesis $H_ {1}$. This indicates that the two coordinate variables are not independent of each other. That is, the coordinate changes of the pixels in the set $D$ show regularity. And these points can form the edges of the image. If $p>\alpha$, we accept the null hypothesis $H_ {0}$ and assume that the two coordinate variables are independent of each other. This means that the variation between the two point coordinates is arbitrary. Therefore, these points cannot form the edges of the image.

\begin{table}[h]
	\centering
	\caption{Fisher's test contingency table for continuous edge images}
	\label{lianxu1}
	\begin{tabular}{c  c c c }
		\toprule 
		& $\left |  \Delta y \right |\le k$ & $\left |  \Delta y \right |> k$ & total\\
		\midrule 
		$\left |  \Delta x \right |\le  k$ & 83 & 0  & 83\\
		$\left |  \Delta x \right |> k$ & 89 & 10 & 99 \\
		total & 172 & 10 & 182 \\
		\bottomrule 
	\end{tabular}
\end{table}

\begin{table}
	\begin{center}
		\caption{Chi-square test contingency table for discontinuous edge images}
		\label{lisan1}
		\begin{tabular}{c  c c c }
			\toprule
			& $\left |  \Delta y \right |\le k$ & $\left |  \Delta y \right |> k$ & total\\
			\midrule
			$\left |  \Delta x \right |\le  k$ & 21 & 24  & 45\\
			$\left |  \Delta x \right |> k$ & 21 & 15  & 36\\
			total & 42 & 39 & 81 \\
			\bottomrule
		\end{tabular}
	\end{center}
\end{table}

In the EDIT method, the red line segment in Fig. \ref{dlexample}(a) marks a continuous curve. Select the area where the red line segment is located in the red box, and then divide the area into $11\times 10 $ regions $\left \{ R_{k}  \right \} _{k=1}^{11\times 10}  $. Figure 8 (b) shows a window with a smooth edge curve and the regions where the pixels are located are marked in the Fig. \ref{dlexample}(b) with coordinates. And Fig. \ref{dlexample}(c) shows a window with noise. The lower left corner is set to (0,0), the horizontal axis is right, and the vertical axis is up.

In Fig. \ref{dlexample}(b), all pixel coordinates form the set $D$, and the $x, y$ coordinates of the pixels are the two categorical variables. For example, the coordinates of point $A$ are (2, 3), and the coordinates of point $B$ are (1, 1). Therefore, the relative displacement of point $A$ with respect to point $B$ in the horizontal direction is $|\Delta x|=|x_{1}-x_{2}|=|2-1|=1$, the relative displacement along the vertical axis is $|\Delta y|= |y_{1}-y_{2}|=2$. Similarly, the relative displacement among all points can be calculated. Then, we set the relative displacement limit $k=3$, and count the frequency of $|\Delta x|, |\Delta y|$ occurring inside and outside the limit $k$. After the above operation, the contingency Table \ref{lianxu1} can be obtained. Observing the frequency in the table, it is found that when $\left |  \Delta y \right |> k$ and $\left |  \Delta x \right |\le  k$, the frequency is less than 5. Therefore, Fisher's exact test is used to determine whether the two coordinate variables are independent of each other.

We can calculate $p=0.0021$ according to the formula \eqref{fisherbz}, which is much smaller than the significance level $\alpha=0.05$. Therefore, we reject the null hypothesis $H_ {0}$ and accept the alternative hypothesis $H_ {1}$. This indicates that the $x, y$ coordinates of pixels in the image are not independent of each other. The changes in these pixels are regular, and they can make up the edges of an image. Thus, we determine the pixels in Fig. 8(a) as edge points.

To verify whether the above method is effective on images containing noise points, we will continue to perform the test on Fig. \ref{dlexample}(c). Based on the same statistical method as shown in Fig. \ref{dlexample}(b), we obtained a contingency table for two categorical variables $\Delta x, \Delta y$, as shown in Table \ref{lisan1}. It was observed that the frequencies in the table were all greater than 5, so a chi-square test was used. After calculation, the test result shows that $p=0.3722$, which is higher than the significance level. Therefore, we accepting the null hypothesis $H_ {0}$, the $x$ and $y$ coordinates of these points are independent of each other. It means that the coordinate changes of the pixels in this area are irregular and cannot form a continuous edge. In summary, we can demonstrate the effectiveness of the proposed method. See more details in Algorithm  \ref{alg:CAM-EDIT}.

\section{Experiment}
In order to verify the usability of this method, this algorithm is applied to image edge detection task. This section is divided into three parts. First, we introduce the public data set and edge evaluation indicators used in the experiment. Then, the choice of parameters is discussed. Finally, we compare the performance of our method with other algorithms (ED\cite{Cihan2012Edge}, PEL\cite {Akinlar0PEL}, CannySR\cite{Akinlar2015} , CannySRPF, EDPF\cite {2012EDPF}, TIP2020\cite {2020An}, MSCNGP\cite {2024Edge}) by MSE (mean square error), PSNR (peak signal to noise ratio) and F1 fraction.

\begin{algorithm}[H]
	\caption{Channel Attention Mechanism-Edge Detection based on Independence Test}\label{alg:CAM-EDIT}
	\begin{algorithmic}[1]
		\STATE {\textbf{Input:}} Input\_image, filter kernel $kernel\_size=5$, contingency table $a=b=c=d=0$, significance level $\alpha=0.05$, displacement limit $k=3$
		\STATE {\textbf{Output:}} Result\_image
		\STATE Define the channel attention model to extract edge features;
		\STATE Calculate and normalize gradient values;
		\STATE Partition gradients using the sigmoid function into membership degrees;
		\STATE Apply a median filter ('kernel=5') and morphological operations for coarse graining;
		\STATE Divide image into $M \times N$ regions ($conditions$);
		\FOR{each region}
		\STATE Define a $M \times N$ window in the region and extract the image;
		\FOR{$(x_{1}, y_{1}) \in D$ and $(x_{2}, y_{2}) \in D$}
		\STATE Compute $\left | \bigtriangleup x \right |$ and $\left | \bigtriangleup y \right |$;
		\IF{$\left | \Delta x \right | \le k$ and $\left | \Delta y \right | \le k$}
		\STATE $a += 1$;
		\ELSIF{$\left | \Delta x \right | \le k$ and $\left | \Delta y \right | > k$}
		\STATE $b += 1$;
		\ELSIF{$\left | \Delta x \right | > k$ and $\left | \Delta y \right | \le k$}
		\STATE $c += 1$;
		\ELSIF{$\left | \Delta x \right | > k$ and $\left | \Delta y \right | > k$}
		\STATE $d += 1$;
		\ENDIF
		\ENDFOR
		\STATE Move the window by 5 pixels to the right, and shift down by 5 pixels when the window reaches the far right of the region.
		\IF{$any$ $of \{ a, b, c, d \} < 5$}
		\STATE Perform Fisher's exact test and filter based on the $p$ value.
		\ENDIF
		\IF{$all$ $of \{ a, b, c, d \} > 5$}
		\STATE Perform chi-square test and filter based on the $p$ value.
		\ENDIF
		\ENDFOR
		\STATE \textbf{return} Final edge image using the pixels within the filtered region.
	\end{algorithmic}
\end{algorithm}

\subsection{Dataset and edge evaluation metrics}

\subsubsection{Datasets}
We evaluate the performance of our CAM-EDIT on BSDS500\cite{bsds500} and NYUDv2\cite{Silberman2012Indoor} edge datasets. The BSDS500 \cite{bsds500} is a publicly available benchmark for image segmentation, created by UC Berkeley's Computer Vision Lab. It contains 500 images, each annotated for edges by 4-9 annotators. Widely used in edge detection tasks, the dataset is divided into training, validation, and testing sets to support algorithm development and evaluation.The NYUDv2 dataset\cite{Silberman2012Indoor} consists of 1449 RGB D indoor scene images, each with a resolution of 480 × 640 pixels. This dataset was originally designed for scene understanding and has been widely used to evaluate edge detection performance in recent years. In this paper, according to the method of Liu et al. \cite{Liu2017RCF}, we divide the dataset into 869 training samples and 580 test samples.

\subsubsection{Evaluation metrics}
We calculated the MSE, PSNR and F-measure value of images processed by different algorithms. MSE represents the similarity between the input and standard images, with a lower MSE indicating better edge detection. The PSNR values further help assess the quality of edge preservation in the processed images. The calculation formula is as follows:
\begin{equation}
	MSE=\frac{1}{MN}\sum_{i=1}^{M}\sum_{j=1}^{N}[f^{'}(i,j)-f(i,j) ]^{2}  
	\label{mse}
\end{equation}
where $f^{'}(i,j)$ and $f(i,j)$ are the detected edge image and the original input image, respectively, while $M$ and $N$ are the width and height of the image.

PSNR quantifies the degree of image distortion by calculating the ratio of the peak value of the signal to the noise. It is currently another widely used indicator for popularizing image edges. The larger the PSNR value, the smaller the difference between images, and the more accurate the detected image edges. Its calculation formula is as follows: 
\begin{equation}
	PSNR=10log_{10}(\frac{MN\times (\text{max}(L))^{2} }{ {\textstyle \sum_{i=1}^{M} {\textstyle \sum_{j=1}^{N}[f^{'}(i,j )-f(i,j)]^{2} } } } ) 
	\label{psnr}
\end{equation}
where the max is the maximum possible value of the image pixel, and $L$ is the number of bits per sampling point. Generally, the sampling point is 8 bits, i.e. $L=8$. Then max($L$)=255 represents the maximum brightness value of the image.

Following existing methods\cite{Pu2022Edter}, the F-measure value is frequently used for comparing the performance of edge detection methods. The calculation formula is as follows:
\begin{equation}
	F=\frac{2\times Presion\times Recall}{Presion+Recall}  
	\label{Fmeasure}
\end{equation}
where precision (\(\text{TP} / (\text{TP} + \text{FP})\)) represents the likelihood that an edge pixel predicted by the model is indeed a true edge pixel. Recall (\(\text{TP} / (\text{TP} + \text{FN})\)) measures the proportion of actual edge pixels that are correctly detected. Here, TP, TN, FP, and FN denote the number of true positives, true negatives, false positives, and false negatives, respectively.

\subsection{Parameter selection}

\begin{figure}[]
	\centering
	\includegraphics[width=2.5in]{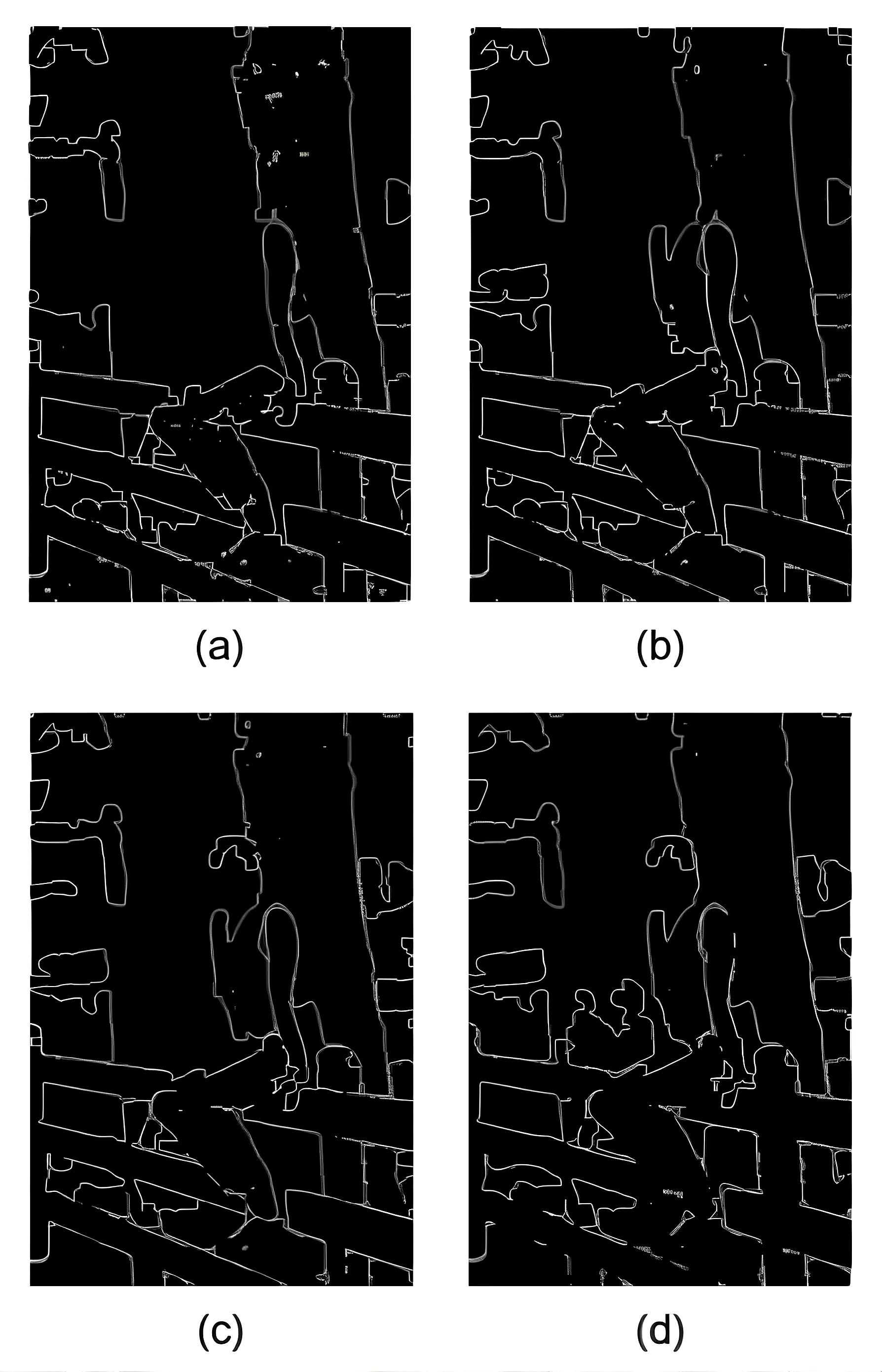}
	\caption{The edge detection results of kernels of different sizes in the median filter. (a) $kernel=1$; (b)  $kernel=3$; (c) $kernel=5$; (d) $kernel=7$.}
	\label{kernel}
\end{figure}

We use a median filter for coarse graining. Smaller kernels preserve details and sharp edges but may not remove noise effectively. Larger kernels reduce noise but can blur details and edges. We tested different kernel sizes for edge detection, calculating the MSE and PSNR for each, as shown in Fig. \ref{kernel} and Table \ref{kernel1}. The kernel of size 1 simulates the effect without using a filter.
	
\begin{table}
	\begin{center}
		\caption{PERFORMANCE OF DIFFERENT FILTER PARAMETERS IN CAM-EDIT}
		\label{kernel1}
		\begin{tabular}{c  c c c c}
			\toprule
			Method & $kernel=1$ & $kernel=3$ & $kernel=5$ & $kernel=7$ \\
			\midrule
			MSE \textdownarrow & 5.7093 &  5.5832 & 5.5238 & \textbf{5.4799}  \\
			PSNR \textuparrow & 40.5649 & 40.6620 & 40.7084 & \textbf{40.7431} \\
			\bottomrule
		\end{tabular}
	\end{center}
	\footnotesize {The bold represents the best result.}
\end{table}

The results show that when the median filter core size is set to 7, MSE and PSNR achieve the best performance, followed by the planting filter with the size of 5. However, the visual effect of a kernel size of 7 is not as good as that of a kernel size of 5. Therefore, we select a core size of 5 for the coarsening of the median filter.

\subsection{Comparison of different algorithms}
\begin{figure}[]
	\centering
	\includegraphics[width=3.5in]{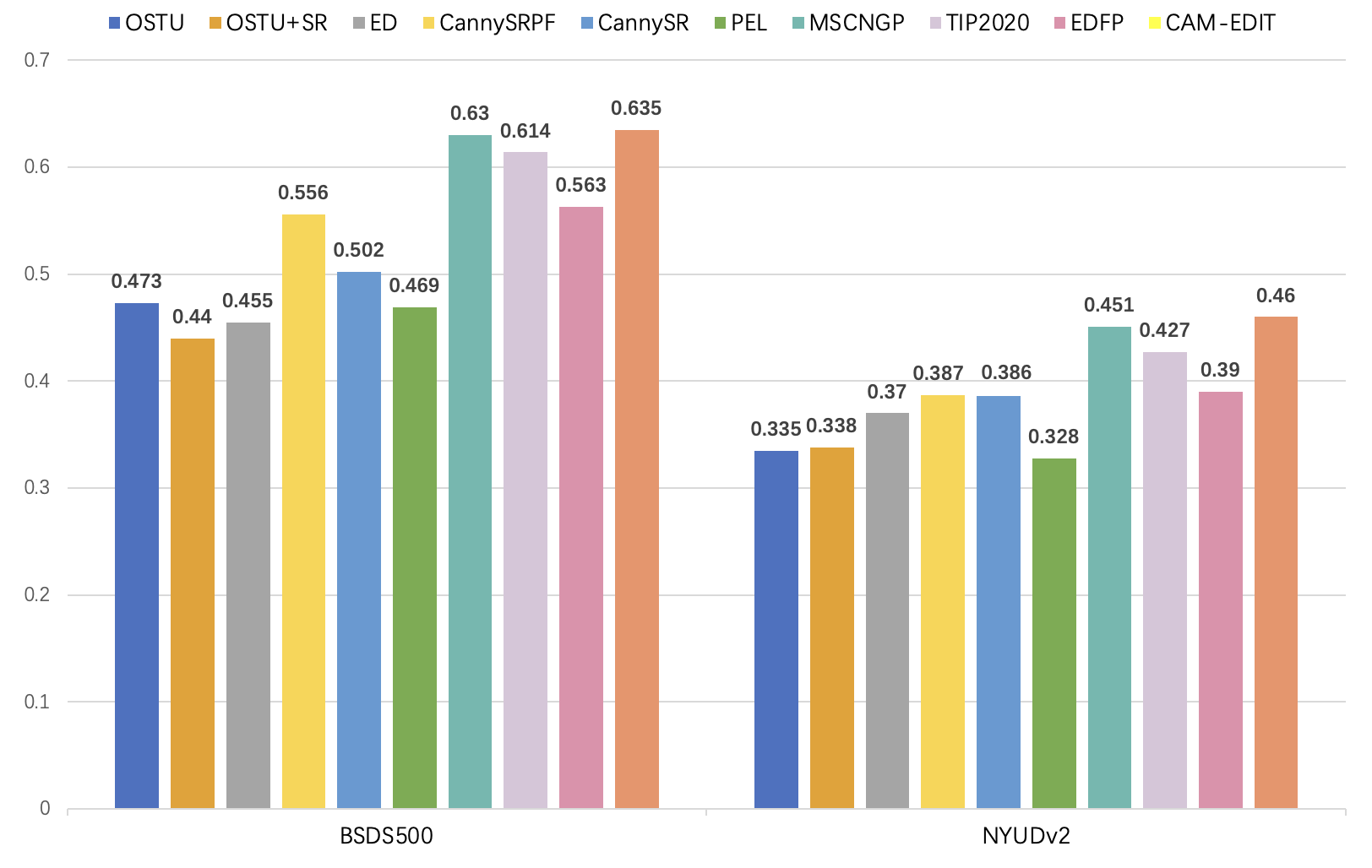}
	\caption{F-measure of the nine methods on both BSDS500 (left half) and NYUDv2 (right half). The nine colors represent the nine methods, respectively.}
	\label{shujuji}
\end{figure}

The results of F-measure on the whole testing images of both BSDS500 and NYUDv2 are shown in Fig. \ref{shujuji} and Table \ref{F}. It shows that the CAM-EDIT method achieves 0.635 and 0.460 for BSDS500 and NYUDv2, respectively. The MSCNGP and TIP2020 methods followed with second best scores of 0.630 and 0.614 on the BSDS500. This demonstrates our algorithm's high accuracy and robustness in detecting edges of various shapes, sizes, and directions in complex scenes.
﻿

\begin{table}
	\begin{center}
		\caption{The F-measure of different algorithms on the BSDS500}
		\label{F}
		\begin{tabular}{c c c}
			\toprule
			& BSDS500 & NYUDv2 \\
			\midrule
			OSTU & 0.473  & 0.335\\
			OSTU+SR & 0.440 & 0.338 \\
			ED \cite{Cihan2012Edge} & 0.455 & 0.370 \\
			PEL \cite{Akinlar0PEL} & 0.469 & 0.328 \\
			CannySR \cite{Akinlar2015} & 0.502 & 0.386 \\
			CannySRPF & 0.556 & 0.387\\
			EDPF \cite{2012EDPF} & 0.563 & 0.390\\
			TIP2020 \cite{2020An} & 0.614 & 0.427\\
			MSCNGP \cite{2024Edge} & 0.630 & 0.451\\
			CAM-EDIT & \textbf{0.635} & \textbf{0.460}\\
			\bottomrule
		\end{tabular}
	\end{center}
	\footnotesize {The bold represents the maximum F-measure score.}
\end{table}

\begin{table}
	\renewcommand{\arraystretch}{1} 
	\begin{center}
		\caption{Evaluation results of images processed by different algorithms}
		\label{butongsuanfa}
		\begin{tabular}{c c c c c c }
			\toprule
			& image &  Canny &  CannySR &  TIP2020 &  CAM-EDIT \\
			\midrule
			&  Image 1 & 5.532 & 5.501 & 4.982 & \textbf{4.653} \\
			& Image 2 & 9.932 & 9.876 & 8.759 & \textbf{7.984} \\
			MSE \textdownarrow & Image 3 & 10.786 & 10.779 & 9.808 & \textbf{9.255} \\
			& Image 4 & 6.141 & 6.114 & 5.159 &  \textbf{4.969} \\
			& Image 5 & 12.806 & 12.851 &  11.871 &  \textbf{10.989} \\
			\midrule
			& Image 1 & 40.702  & 40.726 & 41.362 & \textbf{41.454} \\
			& Image 2 & 38.161 & 38.185 & 38.236 & \textbf{39.109} \\
			PSNR \textuparrow & Image 3  & 37.802 & 37.805 & 37.973 & \textbf{38.467} \\
			& Image 4 &  40.284 & 40.267 & 40.360 & \textbf{41.169} \\
			& Image 5 &  37.057 & 37.042 & 37.345 & \textbf{37.721} \\
			\bottomrule
		\end{tabular}
	\end{center}
	\footnotesize {The bold represents the best result.}
\end{table}

\begin{figure*}[htbp]
	\centering
	\includegraphics[width=7in]{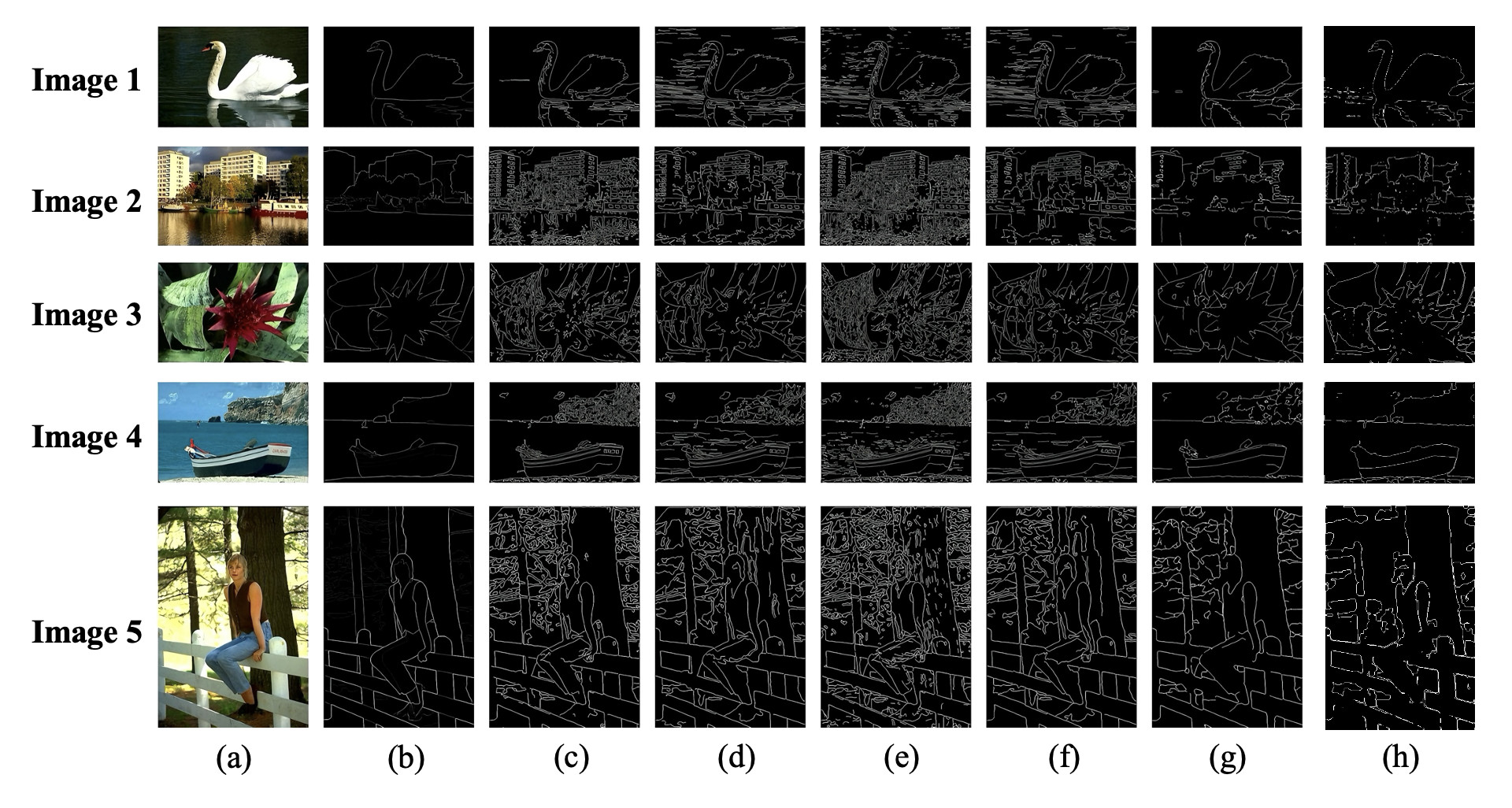}
	\caption{Edge Detection Results on BSDS500. (a) Original images. (b) Ground truths. (c) CannySR. (d) CannySRPF. (e) ED. (f) EDPF. (g) MSCNOGP. (h) CAM-EDIT.}
	\label{BSDS500}
\end{figure*}

We use BSDS500 dataset and NYUDv2 dataset to compare the proposed algorithm with five popular algorithms. The Fig. \ref{BSDS500}  shows the edge detection results of five randomly selected images in the BSDS500 dataset. The first three algorithms capture too much detail and noise at the same time of edge detection, resulting in low accuracy. ED and EDFP algorithms perform better, producing clearer and more continuous edges. On the contrary, our algorithm further reduces noise interference, retains important edge information, and provides better visual effect.

Table \ref{butongsuanfa} compares the MSE and PSNR of different edge detection algorithms on the five images mentioned above. In these five images, our algorithm shows the lowest MSE, indicating the best edge accuracy and integrity, followed by the TIP2020 algorithm. The traditional Canny and CannySR algorithms have higher MSE values. For PSNR, our algorithm achievs the highest value, displaying smoother edges and less noise, while the traditional Canny algorithm had the lowest PSNR.

Furthermore, we conducted a comparative analysis of the F-measure, MSE, and PSNR metrics on the BSDS500 and NYUDv2 datasets across three benchmark algorithms: Canny, CannySR, and TIP2020. The results, presented in Table \ref{quantitative} , clearly demonstrate that CAM-EDIT achieves superior performance.

\begin{table}
	\begin{center}
		\caption{Evaluation results of different algorithms on the entire dataset}
		\label{quantitative}
		\begin{tabular}{c  c c c c}
			\toprule
			Method & BSDS500(F1) & NYUDv2(F1) & PSNR \textuparrow & MSE \textdownarrow \\
			\midrule
			Canny & 0.501 & 0.386 & 37.05 & 12.81  \\
			CannySR & 0.502& 0.386 & 37.03 & 12.85 \\
			TIP2020 & 0.614 & 0.427 & 39.36 & 9.81 \\
			CAM-EDIT & \textbf{0.635} & \textbf{0.460} & \textbf{40.33} & \textbf{7.03} \\
			\bottomrule
		\end{tabular}
	\end{center}
	\footnotesize {The bold represents the best result.}
\end{table}

\subsection{Robustness analysis of the CAM-EDIT}

\begin{figure}[]
	\centering
	\includegraphics[width=2.5in]{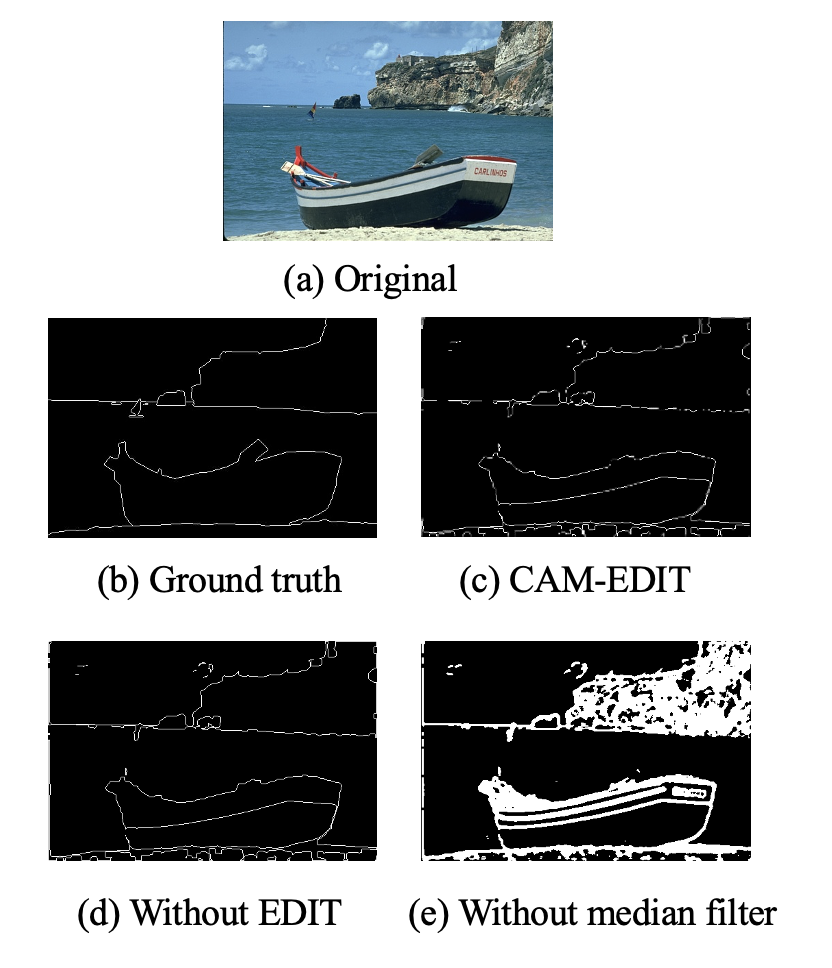}
	\caption{The edge is generated under three different noise conditions with CAM-EDIT. (a) Original. (b) Ground Truth. (c) CAM-EDIT. (d) without EDIT. (e) without median filter.}
	
	\label{bu1}
\end{figure}

\begin{table*}
	\renewcommand{\arraystretch}{1} 
	\begin{center}
		\caption{PERFORMANCE OF CAM-EDIT WITHOUT EDIT AND WITHOUT MEDIAN FILTER IN EDGE DETECTION}
		\label{shiyanduibi1}
		\begin{tabular}{c c c c c c c c c c c}
			\toprule
			& & & MSE \textdownarrow & & & & & PSNR \textuparrow & & \\
			\cmidrule{2-11} 
			image & Image 1 & Image 2 & Image 3 & Image 4 & Image 5 & Image 1 & Image 2 & Image 3 & Image 4 & Image 5 \\
			\midrule
			without EDIT & 3.3222 & 7.0325 & 6.2225 & 2.6001 & 6.7847 & 42.9166 & 40.3258 & 40.1912 & 43.9795 & 39.8155 \\
			without median filter & 3.5075 & 7.5817 & 6.5610 & 2.9132 & 7.2118 & 42.6808 & 39.9474 & 39.9611 & 43.4871 & 39.5504 \\
			CAM-EDIT & \textbf{3.3202} & \textbf{7.0208} & \textbf{6.2195} & \textbf{2.5936} & \textbf{6.7829} & \textbf{42.9192} & \textbf{40.3342} & \textbf{40.1932} & \textbf{43.9918} & \textbf{39.8166} \\
			\bottomrule
		\end{tabular}
	\end{center}
	\footnotesize {The bold represents the best result.}
\end{table*}

This section validates the necessity of the coarse-grained operation and independence check incorporated into the proposed algorithm. First, we analyze the edge detection performance of CAM-EDIT with and without the coarse-graining operation, as illustrated in Fig. \ref{bu1} (d). It is observed that, in the absence of coarse graining, a substantial number of non-edge pixels appear along the image boundaries, accompanied by noticeable noise artifacts within the image. Second, when the EDIT step is omitted, although the resulting image contains fewer noise points, the detected edges are significantly thicker than the standard ones, compromising edge localization accuracy.

To further substantiate these observations, we compute the MSE and PSNR values for five selected images from the BSDS500 dataset. The result is shown in Table \ref {shiyanduibi1}. Taking the second image as an example, the MSE for the case without the EDIT is 7.0325, with a corresponding PSNR of 40.3258, whereas omitting the coarse-graining operation yields an MSE of 7.5817 and a PSNR of 39.9474. In contrast, CAM-EDIT, which integrates both coarse graining and EDIT, achieves an optimal balance with an MSE of 7.0282 and a PSNR of 40.3342. These quantitative results indicate that the absence of the EDIT step leads to the most suboptimal performance, while the combined application of both steps in CAM-EDIT produces the most favorable outcome.

\begin{figure}[]
	\centering
	\includegraphics[width=2.5in]{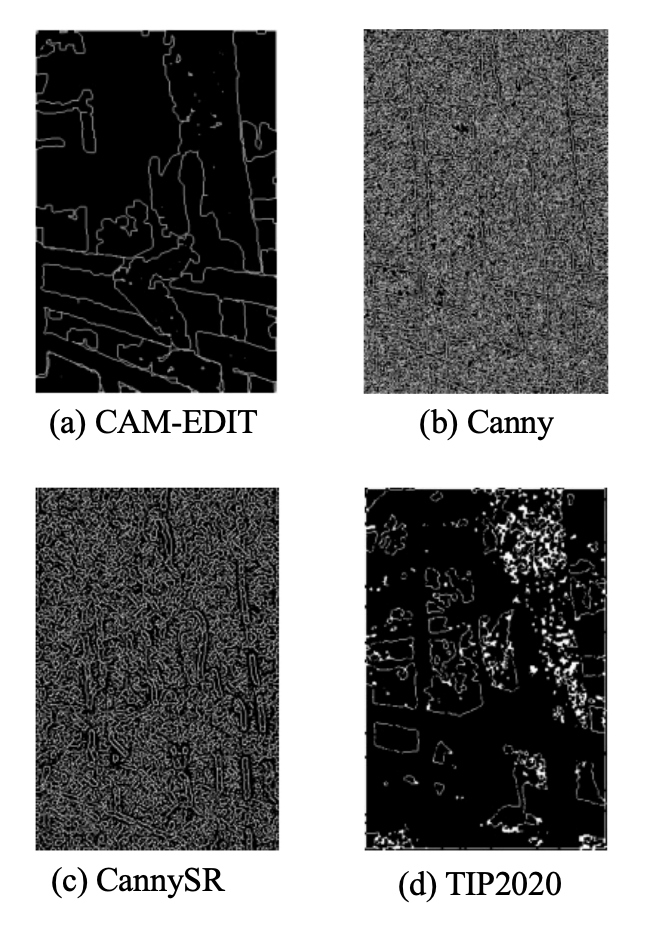}
	\caption{Comparison of edge detection with Gaussian noise.}
	\label{gaosi}
\end{figure}

To test our algorithm's noise immunity, we added Gaussian noise to the images and compared results with previous algorithms. Fig. \ref{gaosi} shows that the traditional Canny algorithm performs the worst, struggling with noise interference and failing to accurately detect edges. While CannySR and TIP2020 reduce noise to some extent, their results still contain many noise points and appear blurry. In contrast, our algorithm still exhibits significant superiority even when it contains a large number of noise points. In addition, we can see from Table \ ref {gaosi1} that the MSE value of CAM-EDIT is 10.963, which is the smallest among the compared algorithms, and the PSNR value of 37.731 is the largest among the compared algorithms. It can be seen that the algorithm proposed in this article has strong performance in noise resistance.

\begin{figure}[]
	\centering
	\includegraphics[width=3.3in]{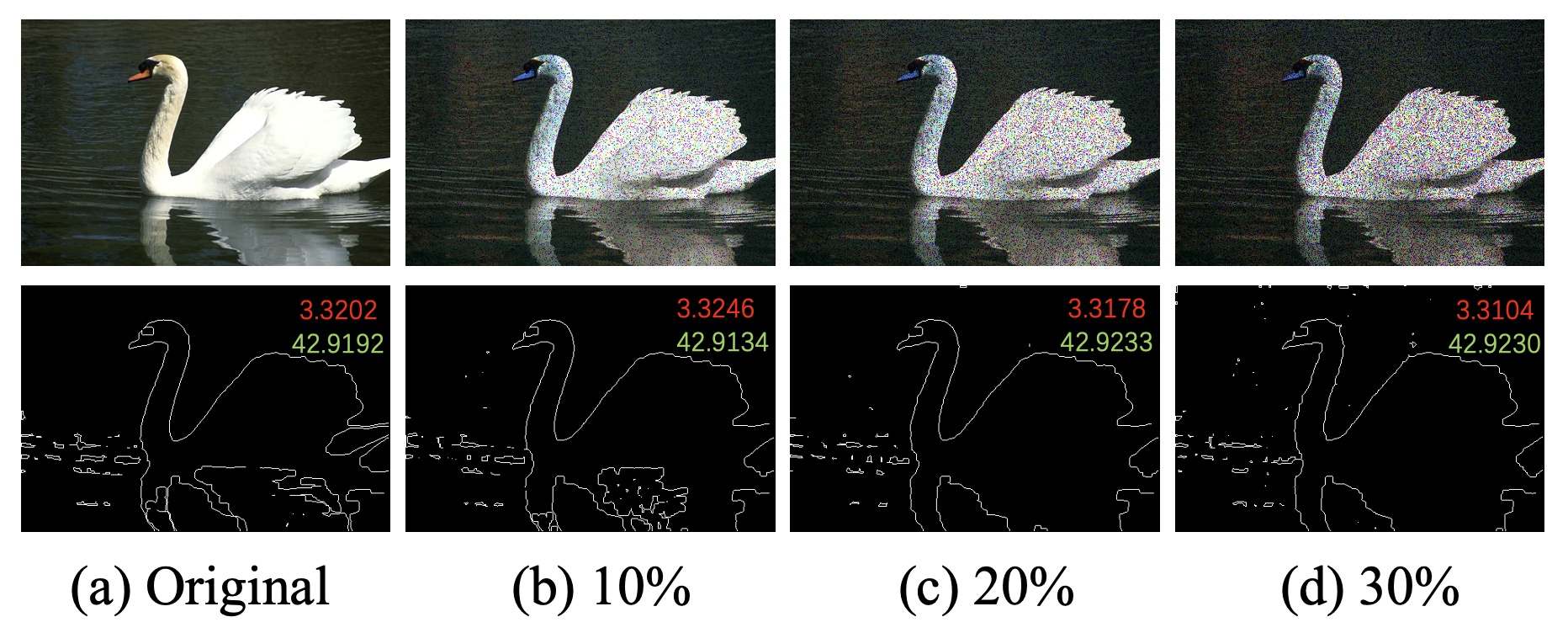}
	\caption{The edge is generated under three different noise conditions with CAM-EDIT. (a) 0\%(original image). (b) 10\% noise. (c) 20\% noise. (d) 30\% noise.}
	
	\label{zao}
\end{figure}

As shown in Fig. \ref{zao} (a) is the original image with different noises; Figure \ref {zao}(b) shows the CAM-EDIT detection results with 10\% noise added; Figure \ref {zao}(c) shows the CAM-EDIT detection results with 20\% noise added; Figure \ref{zao}(d) shows the CAM-EDIT detection results with 30\% noise added. The MSE values are displayed in red font, which are 3.3202, 3.3246, 3.3178, and 3.3140, respectively; The PSNR values are displayed in green font and are 42.9192, 42.9134, 42.9133, and 42.9330, respectively.

\section{Conclusion}

This article presents CAM-EDIT, an edge detection method that combines independence testing and channel attention mechanisms. It enhances accuracy and robustness by removing excessive detail while preserving edge features. The method uutilizes ses channel attention to extract and refine edge features, reduces noise interference, and preserves relevant edge regions through independence checks. Experimental results demonstrate that CAM-EDIT excels in error analysis (MSE and PSNR) and F-measure score, effectively preserving edge information and reducing noise, thereby producing clearer and better visual results compared to traditional algorithms.

The algorithm has the following limitations: (1) It depends on multiple parameters (e.g., channel attention weights, kernel size for dilation and erosion, region partitioning strategy), which requires extensive experimentation and experience, increasing complexity. (2) Noise or complex backgrounds may affect channel attention performance, leading to false or missed detections with similar edge colors. Future work will focus on addressing these issues.

	\bibliographystyle{IEEEtran}
	\bibliography{reference.bib}

\flushend
\end{document}